\documentclass[letterpaper, 10 pt, conference]{ieeeconf}  
\IEEEoverridecommandlockouts                              
\usepackage{cite}
\usepackage{amsmath,amssymb,amsfonts}
\usepackage{algorithmic}
\usepackage{graphicx}
\usepackage{textcomp}
\usepackage[utf8]{inputenc}

\usepackage{lipsum}

\usepackage{xcolor}
\def\hlfon{1} 
\def\todo#1{%
  \ifnum\hlfon=1
    \textcolor{red}{??#1}%
  \else
  \fi
    }

\usepackage[nolist,nohyperlinks]{acronym}
\acrodef{vio}[VIO]{Visual-Inertial Odometry}
\acrodef{rio}[RIO]{Radar-Inertial Odometry}
\acrodef{lio}[LIO]{LiDAR-Inertial Odometry}
\acrodef{ekf}[EKF]{Extended Kalman Filter}
\acrodef{iekf}[IEKF]{Iterated Extended Kalman Filter}
\acrodef{fmcw}[FMCW]{Frequency Modulated Continuous Wave}
\acrodef{imu}[IMU]{Inertial Measurement Unit}
\acrodef{fft}[FFT]{Fast Fourier Transform}
\acrodef{fov}[FoV]{Field of View}
\acrodef{ransac}[RANSAC]{Random Sample Consensus}
\acrodef{dof}[DoF]{Degree of Freedom}
\acrodef{swap}[SWaP]{Size, Weight, and Power}
\acrodef{ape}[APE]{Absolute Pose Error}
\acrodef{rpe}[RPE]{Relative Pose Error}
\acrodef{cfar}[CFAR]{Constant False Alarm Rate}
\acrodef{snr}[SNR]{Signal to Noise Ratio}
\acrodef{soc}[SoC]{System on a Chip}
\acrodef{gnss}[GNSS]{Global Navigation Satellite System}
\acrodef{odr}[ODR]{Orthogonal Distance Regression}

\usepackage{bm}

\usepackage{hyperref}
\usepackage[capitalise]{cleveref}

\usepackage{booktabs}
\usepackage{multirow}
\usepackage{multicol}
\usepackage{makecell}

\usepackage{siunitx}
\usepackage{flushend}

\newcommand{\mat}[1]{\mathbf{#1}}

\renewcommand{\vec}[3]{\bm{#1}_\mathtt{#2}^\mathtt{#3}}
\newcommand{\coord}[1]{$\{\mathtt{#1}\}$}

\newcommand{\Noise}[1]{\bm{n}_{#1}}
\newcommand{\jacobian}[2]{\frac{\partial#1}{\partial#2}}

\newcommand{\rio}{\texttt{RIO}}
\newcommand{\vio}{\texttt{VIO}}
\newcommand{\method}{\texttt{RadVIO}}
\newcommand{\rrxio}{\texttt{RRxIO}}

\newcommand{\forestBase}{Forest}
\newcommand{\indoorsBase}{Indoors}
\newcommand{\fieldBase}{Field}
\newcommand{\forest}{\emph{\forestBase{}}}
\newcommand{\indoors}{\emph{\indoorsBase{}}}
\newcommand{\field}{\emph{\fieldBase{}}}
\newcommand{\Forest}{\textbf{\forestBase{}}}
\newcommand{\Indoors}{\textbf{\indoorsBase{}}}
\newcommand{\Field}{\textbf{\fieldBase{}}}

\newcommand{\experiment}[1]{\texttt{#1}}

\newcommand{\spacingFigureCaption}{\vspace{-4ex}}
\newcommand{\spacingFigureText}{\vspace{-2ex}}
\newcommand{\spacingTableCaption}{\vspace{-1ex}}
\newcommand{\spacingTableText}{\vspace{-2ex}}

\begin{document}

\title{Tightly-Coupled Radar-Visual-Inertial Odometry
\thanks{This work was supported by the Horizon Europe grants SPEAR (101119774) and AUTOASSESS (101120732).}
\thanks{All authors are with the Department of Engineering Cybernetics, O. S. Bragstads Plass 2D, Norwegian University of Science and Technology (NTNU), Trondheim, Norway. Corresponding author: \texttt{morten.nissov@ntnu.no}}
}
    
\author{Morten Nissov, Mohit Singh, and Kostas Alexis}

\maketitle

\begin{abstract}
\ac{vio} is a staple for reliable state estimation on constrained and lightweight platforms due to its versatility and demonstrated performance. However, pertinent challenges regarding robust operation in dark, low-texture, obscured environments complicate the use of such methods. Alternatively, \ac{fmcw} radars, and by extension \ac{rio}, offer robustness to these visual challenges, albeit at the cost of reduced information density and worse long-term accuracy. To address these limitations, this work combines the two in a tightly coupled manner, enabling the resulting method to operate robustly regardless of environmental conditions or trajectory dynamics. The proposed method fuses image features, radar Doppler measurements, and \ac{imu} measurements within an \ac{iekf} in real-time, with radar range data augmenting the visual feature depth initialization. The method is evaluated through flight experiments conducted in both indoor and outdoor environments, as well as through challenges to both exteroceptive modalities (such as darkness, fog, or fast flight), thoroughly demonstrating its robustness. The implementation of the proposed method is available at: \url{https://github.com/ntnu-arl/radvio}.
\end{abstract}

\section{Introduction}
Vision, and by extension \ac{vio}, provides lightweight, low-power sensing, equipping robotic systems with high-density information and reliable odometry. This is especially relevant for \ac{swap} constrained systems that cannot afford to use more accurate sensing solutions, such as LiDARs, but still aim to navigate in environments without \ac{gnss} access. 
Vision has thus become a staple for state estimation on constrained platforms, such as micro aerial vehicles. However, real-world challenges, such as varying illumination conditions, motion blur, and lack of visual texture, complicate the use of vision-based estimation, whereas less common hurdles, such as obscurants, can prohibit its use altogether. 

The desire to enable resilient operation in any condition and environment is, in part, what prompts the newfound research interest into additional modalities, such as \ac{fmcw} radar. Due to the increased wavelength and active sensing, \ac{fmcw} radar does not suffer from the same scenarios that plague typical vision-based systems. However, radars, especially small and compact ones, face their own challenges with noisy, sparse point clouds, which are directly a result of sensor design compromises.
Like \ac{gnss} and \ac{imu} sensors, vision and radar exhibit complementary characteristics, which a fusion methodology can use to alleviate their individual shortcomings. The direct velocity-measurement capabilities of \ac{fmcw} radars yield robust, computationally efficient odometry that is consistently accurate over short time horizons. This provides reliable prior estimates that enable vision to achieve its superior long-term accuracy. However, the fusion needs to be designed carefully in order to best utilize each modality. For example, particularly sparse radar measurements can compromise the least-squares calculation of linear velocity, which many other approaches use, even if only for outlier rejection. 

\begin{figure}[t!]
    \centering
    \includegraphics[width=\linewidth]{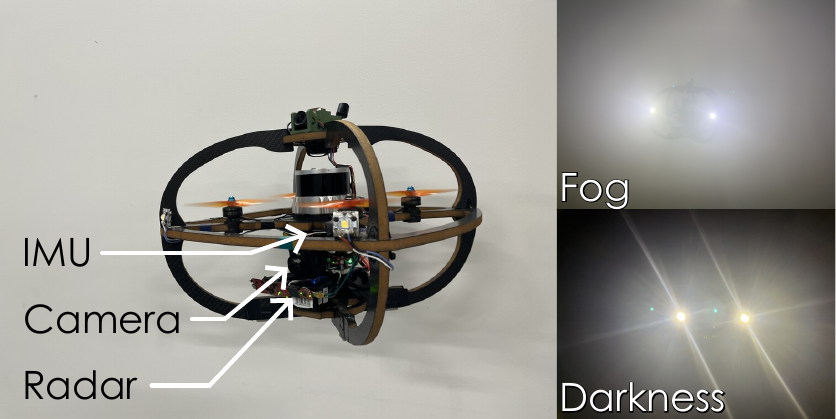}
    \spacingFigureCaption{}
    \caption{The aerial platform used to evaluate the proposed method, shown flying in nominal, obscured (due to fog), and dark environments.}
    \label{fig:intro}
    \spacingFigureText{}
\end{figure}

Motivated by these observations, we propose a tightly-coupled radar-visual-aided inertial navigation system that relies on direct updates of image features (tracked as part of the state vector), Doppler measurement updates from the radar, and image-feature depth initialization from a local window of the aggregated radar point cloud. Thus, our contributions are as follows
\begin{itemize}
    \item Tightly-coupled radar-visual-inertial fusion with extrinsic estimation for radar and vision, enabling robust performance across challenging environments.
    \item A method for image feature depth initialization from sparse and noisy radar point clouds.
    \item Extensive evaluation of the proposed method with flight experiments in challenging conditions for both exteroceptive modalities.
\end{itemize}

The remainder of the paper is organized as follows. Section~\ref{sec:relatedwork} discusses related work, followed by the description of the proposed method in Section~\ref{sec:method}. Evaluation studies are presented in Section~\ref{sec:evaluation}, before conclusions are drawn in Section~\ref{sec:conclusions}.
\section{Related Work}\label{sec:relatedwork}
Since the proposed method lies at the intersection of \ac{vio} and \ac{rio}, we review both individually, followed by existing multi-modal fusion approaches using \ac{fmcw} radar.

\ac{vio} methods have been developed over the years, and many prominent methods have been proposed, which can be separated by their measurement update form (direct or indirect) and by how they track image features (inside or outside of the estimator).
In~\cite{bloesch2017Rovio}, the authors propose a direct method that uses the intensity error between image patches to perform measurement updates. Furthermore, the Multi-State Constraint Kalman Filter~\cite{mourikis2007MSCKF} is a widely acknowledged indirect \ac{vio} method that instead tracks point visual features. Additionally, it proposes multi-state constraint measurement updates to avoid explicitly tracking features in the estimator state vector, thereby reducing computational costs. Furthermore, in~\cite{qin2018vins} the authors develop a windowed optimization framework for tightly-coupled, indirect \ac{vio} and also include a loosely coupled pose graph for handling loop closure. Besides \ac{vio}, there is a rich literature on vision-only state-estimation methods, with ORB-SLAM~\cite{muratal2015orbslam} as a prominent example. A recent, relevant survey covering the applications of vision for estimation is available at~\cite{servieres2021visual}.

Radar-based odometry methods use two main classes of radar sensor: spinning and \ac{soc}. Seminal work of the former category includes \cite{cen2019Matching}, which extracts keypoints from the 2D polar radar image and performs scan-matching for odometry. However, this class of radar sensor is typically large, two-dimensional, and does not usually measure Doppler. As a result, it is not suitable for micro aerial vehicles, which would instead favor the smaller \ac{soc} radar counterpart. Instead, \ac{soc} radars typically return measurements of Doppler as well as range, azimuth, and elevation; however, for a smaller \ac{fov}. In their seminal work \cite{kellner2013instantaneous}, the authors propose a method using \ac{soc} radar to solve a least-squares problem to estimate linear velocity; however, this is limited to 2D. In \cite{doer2020ekf,doer2021yaw}, the authors go on to utilize this least-squares estimate as an aiding source for \ac{rio}, augmenting the method with barometric pressure updates to stabilize height as well as yaw measurement updates, assuming a Manhattan world, to stabilize yaw. More tightly-coupled approaches have also been pursued, for example, in~\cite{kramer2020visually} where the authors propose a factor graph-based estimator with \ac{imu} pre-integration and radial speed factors for estimating velocity and attitude.

In terms of multi-modal fusion, \ac{fmcw} radar is attracting the research community's attention, particularly for its robustness to obscurants and its ability to measure velocity instantaneously, both for fusion with LiDAR and camera sensors. Similar to the aforementioned \ac{rio} works, the authors of~\cite{nissov2024degradation} use a least-squares linear velocity estimate to facilitate LiDAR-radar fusion in environments with obscurants and geometric self-similarities, which are key failures of \ac{lio}. 
In~\cite{nissov2024robust}, the authors fuse radial speed directly to demonstrate how, even with only a single Doppler-bearing pair per measurement, it is feasible to improve \ac{lio} performance in challenging environments. 
The authors of~\cite{noh2025garlio} expand further upon LiDAR-radar fusion by considering a second update stage, after processing LiDAR and radar measurements, with gravity residuals. Considering vision instead, the authors of~\cite{cheng2024msvro} propose a multi-stage fusion methodology for vision and radar, with radar-informed feature selection and joint optimization of visual features and radar linear velocity estimates. In terms of visual-radar-inertial fusion, the most related work~\cite{doer2021Rxio} applies the aforementioned least-squares velocity solution, making use of \ac{odr}, for a loosely-coupled fusion method. While convenient, this can display difficulties in sparse environments, where an accurate solution of the least-squares problem may not be feasible. Furthermore, the lack of extrinsic estimation for the radar can compromise performance, as the calibration routine is not as established as it is for camera-\ac{imu}~\cite{rehder2016Kalibr}. Lastly, the range measurements are not used at all, even though they could otherwise improve visual feature depth initialization, which is a typical problem for monocular camera-based \ac{vio} methods~\cite{sola2012Depth}.
\section{Method}\label{sec:method}

\subsection{Notation and Coordinate Frames}
The manuscript uses coordinate frames to relate measurements from different sensors, including the inertial \coord{I}, body \coord{B}, camera \coord{C}, and radar \coord{R} frames. Body, camera, and radar frames are assumed to be fixed to the platform, such that the transformation between them is rigid and known, approximately, a priori. Furthermore, regarding notation let the quaternion aligning \coord{B} to \coord{I} be $\vec{q}{B}{I}$ (with equivalent rotation matrix $\mat{R}(\vec{q}{B}{I})$) and the translation from \coord{I} to \coord{C}, expressed in \coord{B}, be $\vec{r}{IC}{B}$. Similarly, let the velocity of \coord{C} with respect to \coord{I}, expressed in \coord{B}, be $\vec{v}{IC}{B}$.

\subsection{Basics of \acs{fmcw} Radar}
\ac{fmcw} radars function by transmitting groups of frequency-modulated chirps across an array of antennas, and comparing the transmitted and reflected signals to resolve objects in the environment. By leveraging a two-dimensional \ac{fft}, measurements of range $d$ and radial speed $v_r$ (through the Doppler effect) can be produced. By examining the delay in a reflected signal arriving at different antennas in the array, measurements of azimuth $\theta$ and elevation $\phi$ for a given target in the radar \ac{fov} can be calculated as well. Note that, for the aforementioned measurements, the ability to separate valid returns from noise is critical, for which \ac{cfar} algorithms are typically used \cite{richards2013Fundamentals}. However, this requires sufficient \ac{snr} to distinguish between actual target reflections and noise.

Typical \ac{soc} \ac{fmcw} radars produce point cloud measurements, consisting of targets characterized by position and radial speed within the sensor's \ac{fov}. The position of a target $\bm{p}$ is related to the bearing vector $\bm{\mu}$, which is a function of the azimuth $\theta$ and elevation $\phi$ angles following

\begin{equation}
    \bm{\mu} = \begin{bmatrix} \cos\theta\cos\phi\\ \sin\theta\cos\phi\\ \sin\phi \end{bmatrix},
\end{equation}
such that the position is

\begin{equation}
    \bm{p} = d \bm{\mu}.
\end{equation}
However, individual measurements are typically quite noisy due to effects such as multipath reflections and the limited resolution of the range, Doppler, azimuth, and elevation measurements. As a result, for \ac{fmcw} radar in aided inertial navigation, the radial speed is often of greater interest. Radial speed measurements directly relate to ego-velocity instantaneously, without requiring cross-measurement associations or feature tracking, which are complicated by noise and introduce additional failure points. The radial speed is related to the radar linear velocity $\vec{v}{IR}{R}$ by

\begin{equation}\label{eq:radar:doppler}
    v_r = -\bm{\mu}^\top \vec{v}{IR}{R},
\end{equation}
resulting in a measurement which can be related to the body-frame linear velocity through rigid-frame transforms. Other approaches will estimate $\vec{v}{IR}{R}$ from multiple radar measurements using least squares, typically with \ac{ransac} or M-estimators for added robustness. This is not always feasible as the radar point cloud returns are often very sparse (e.g., due to environmental conditions), sometimes yielding as few as one or two points. The least-squares solution needs at least three points (four to estimate covariance) with sufficient diversity across the \ac{fov} to lend full observability. A more robust approach would use the individual radial speed measurements, as this eliminates the need to explicitly handle such complications.

\subsection{Aided Inertial Navigation}
The proposed method builds on the \ac{vio} implementation proposed in~\cite{bloesch2017Rovio}. This method, known as ROVIO~\cite{bloesch2017Rovio}, is an \ac{iekf}-based estimator that uses FAST corners~\cite{rosten2006Fast} to generate image features across different levels of the image pyramid. Multilevel patches around the features are maintained, and the features are added to the state vector for tracking. Measurement updates can thus be performed using the patch intensity for the innovation term.

This method performs well overall, yet it is naturally subject to the typical limitations of vision-based navigation (i.e., low-light, obscured, or otherwise featureless environments). Thus, there is potential to expand the operating regime by the proposed fusion of radar measurements. Furthermore, this has the dual effect of addressing persistent issues associated with \ac{rio}, namely vertical drift~\cite{nissov2024degradation}, which has traditionally been resolved by incorporating barometers~\cite{doer2021xrio}. However, barometers are unreliable due to imperfect knowledge of the environmental conditions necessary for relating pressure to altitude, as well as ample opportunities for disturbances, both indoors and outdoors~\cite{parviainen2008barometry}.

Key to the proposed method is the robocentric formulation of the dynamics, which reduces linearization errors, as shown in~\cite{castellanos2004Limits}. Additionally, in the robocentric formulation, the long-term drift in the yaw estimate (inevitable due to insufficient observability) does not affect the body-frame velocity estimation, which is critical for control applications. Furthermore, this provides a natural extension to radar measurements, as the radar effectively serves as a body-velocity sensor, following \cref{eq:radar:doppler}. Thus, the estimator state space is composed of the navigation states $\mat{x}_{\mathcal{N}}$ (including position $\vec{r}{IB}{B}$, velocity $\vec{v}{IB}{B}$, and attitude $\vec{q}{B}{I}$)

\begin{equation}
    \mat{x}_{\mathcal{N}} = \begin{pmatrix} \vec{r}{IB}{B}  &\vec{v}{IB}{B} &\vec{q}{B}{I} \end{pmatrix},
\end{equation}
calibration states $\mat{x}_{\mathcal{K}}$ (including accelerometer and gyroscope biases $\vec{b}{a}{}$, $\vec{b}{g}{}$ and the extrinsic transforms between \ac{imu} and camera $\vec{r}{BC}{B}$, $\vec{q}{B}{C}$ and between \ac{imu} and radar $\vec{r}{BR}{B}$, $\vec{q}{B}{R}$)

\begin{equation}
    \mat{x}_{\mathcal{K}} = \begin{pmatrix} \vec{b}{a}{}    &\vec{b}{g}{}   &\vec{r}{BC}{B} &\vec{q}{B}{C}  &\vec{r}{BR}{B} &\vec{q}{B}{R} \end{pmatrix},
\end{equation}
and image feature states $\mat{x}_{\mathcal{F}}$ (including per-feature bearing vector $\bm{\mu}_{i}^{\mathcal{C}}$ and inverse depth parameter $\rho_{i}^{\mathcal{C}}$ for $N_{\mathcal{F}}$ features where $i \in \{ 1, \ldots, N_{\mathcal{F}} \}$) 

\begin{equation}
    \mat{x}_{\mathcal{F}} = \begin{pmatrix} \bm{\mu}_{1}^{\mathcal{C}}  &\ldots &\bm{\mu}_{N_{\mathcal{F}}}^{\mathcal{C}}   &\rho_{1}^{\mathcal{C}} &\ldots &\rho_{N_{\mathcal{F}}}^{\mathcal{C}} \end{pmatrix},
\end{equation}
such that the total state vector $\mat{x}$ is given by

\begin{equation}
    \mat{x} = \begin{pmatrix} \mat{x}_{\mathcal{N}} &\mat{x}_{\mathcal{K}} &\mat{x}_{\mathcal{F}} \end{pmatrix}.
\end{equation}
The dynamics take the specific force $\vec{f}{IB}{B}$ and angular velocity $\vec{\omega}{IB}{B}$ as input, in typical aided inertial navigation fashion, which can be sourced as measurements $\vec{\tilde{f}}{IB}{B}$, $\vec{\tilde{\omega}}{IB}{B}$ from the \ac{imu}. The \ac{imu} measurements are assumed to be corrupted by bias offsets as well as Gaussian-distributed noise, such that the noisy estimates for specific force $\bm{\hat{f}}$ and angular velocity $\bm{\hat{\omega}}$ are given by

\begin{align}
    \bm{\hat{f}} &= \overbrace{(\vec{f}{IB}{B} + \vec{b}{a}{} + \Noise{\vec{f}{IB}{B}})}^{\vec{\tilde{f}}{IB}{B}} - \vec{\hat{b}}{a}{},\\
    \bm{\hat{\omega}} &= \underbrace{(\vec{\omega}{IB}{B} + \vec{b}{g}{} + \Noise{\vec{\omega}{IB}{B}})}_{\vec{\tilde{\omega}}{IB}{B}} - \vec{\hat{b}}{g}{},
\end{align}
for the measurement noises $\Noise{\vec{f}{IB}{B}}$, $\Noise{\vec{\omega}{IB}{B}}$ and the estimates of the accelerometer and gyroscope biases $\vec{\hat{b}}{a}{}$, $\vec{\hat{b}}{g}{}$.
The dynamics of the state space, utilizing the robocentric formulation from~\cite{castellanos2004Limits,bloesch2017Rovio}, can thus be written as follows

\begin{align}
    \vec{\dot{r}}{IB}{B} &= -\bm{\hat{\omega}}^{\times} \vec{r}{IB}{B} + \vec{v}{IB}{B} + \Noise{\vec{r}{IB}{B}},\\
    \vec{\dot{v}}{IB}{B} &= -\bm{\hat{\omega}}^{\times} \vec{v}{IB}{B} + \bm{\hat{f}} + (\vec{q}{B}{I})^{-1} \cdot \bm{g},\\
    \vec{\dot{q}}{B}{I} &= \frac{1}{2} \vec{q}{B}{I} \otimes \bm{\hat{\omega}},\\
    \vec{\dot{b}}{a}{} &= \Noise{\vec{b}{a}{}},\\
    \vec{\dot{b}}{g}{} &= \Noise{\vec{b}{g}{}},\\
    \vec{\dot{r}}{BC}{B} &= \Noise{\vec{r}{BC}{B}},\\
    \vec{\dot{q}}{B}{C} &= \Noise{\vec{q}{B}{C}},\\
    \vec{\dot{r}}{BR}{B} &= \Noise{\vec{r}{BR}{B}},\\
    \vec{\dot{q}}{B}{R} &= \Noise{\vec{q}{B}{R}},
\end{align}
for the white, Gaussian noise sources $\Noise{*}$ and inertial frame gravity $\bm{g}$, and where $(\cdot)^{\times}$ denotes the skew symmetric matrix, $\vec{q}{B}{I} \cdot \bm{x}$ denotes the quaternion rotating an arbitrary vector $\bm{x}\in\mathbb{R}^3$, and $\otimes$ denotes the quaternion product. Feature dynamics are omitted, as they are unchanged from the original work and described in \cite{bloesch2017Rovio}. With respect to the prediction step, to reduce computational cost, ROVIO~\cite{bloesch2017Rovio} propagates the mean and covariance using the average of the \ac{imu} measurements between successive measurement update timestamps. This assumes measurement updates happen sufficiently frequently such that this does not result in excessive errors. 

\begin{figure}[h]
    \centering
    \includegraphics[width=\linewidth]{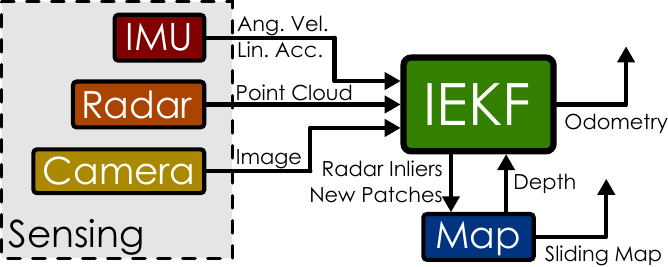}
    \spacingFigureCaption{}
    \caption{Information flow of the proposed method from the \ac{imu}, radar, and camera sensors to the odometry and local-map outputs.}
    \label{fig:method}
    \spacingFigureText{}
\end{figure}

With respect to the measurement updates, image measurements are processed, and the state estimate is updated according to \cite{bloesch2017Rovio} by evaluating the intensities of the tracked feature patches. The radar measurements are used in two ways: first, in a tightly-coupled measurement update, the radial speed for each point in a point cloud is fused. Secondly, a sliding-window voxel map is created from the radar points and used to initialize the depth of newly detected visual features.
The overall structure and information flow of the proposed method are shown in \cref{fig:method}.

\subsubsection{Radial Speed Measurement Update}
From \cref{eq:radar:doppler}, the radial speed of a target from the radar point cloud is related to the radar sensor's ego-velocity. Thus, assuming a static environment, this can be related to the body-frame linear velocity and used as a measurement update. While the point cloud measurement may contain multiple targets, the measurement update is formulated identically for each. As a result, no per-point index is considered here, and the formulation will be presented for an arbitrary target. The same calculations are repeated for each target in the implementation.

The radial speed measurement function $h_{v_r}(\mat{x})$ can be written as

\begin{equation}
    h_{v_r}(\mat{x}) = -\bm{\tilde{\mu}}^\top \left( \vec{q}{B}{R} \cdot \left( \vec{v}{IB}{B} + (\bm{\bar{\omega}} - \vec{b}{g}{})^{\times} \vec{r}{BR}{B} \right) \right),
\end{equation}
where $\bm{\bar{\omega}}$ is the angular velocity interpolated to the middle of the radar chirping period from \ac{imu} measurements and $\bm{\tilde{\mu}}$ is the target's bearing measurement. As a result, the Doppler innovation $e_{\mathcal{D}}$ can be written as

\begin{equation}
    e_{\mathcal{D}} = h_{v_r}(\mat{x}) - \tilde{v}_{r},
\end{equation}
where $\tilde{v}_r$ is the radial speed measurement, modeled with additive, normally-distributed measurement noise with standard deviation $\sigma_{v_r}$. The measurement equation has the following non-zero partial derivatives with respect to the state space

\begin{align}
    \jacobian{h_{v_r}}{\vec{v}{IB}{B}} &= -\bm{\tilde{\mu}}^\top \mat{R}(\vec{q}{B}{R}),\\
    \jacobian{h_{v_r}}{\vec{b}{g}{}} &= -\bm{\tilde{\mu}}^\top \mat{R}(\vec{q}{B}{R}) (\vec{r}{BR}{B})^{\times},\\
    \jacobian{h_{v_r}}{\vec{r}{BR}{B}} &= -\bm{\tilde{\mu}}^\top \mat{R}(\vec{q}{B}{R}) (\bm{\bar{\omega}} - \vec{b}{g}{})^{\times},\\
    \jacobian{h_{v_r}}{\vec{q}{B}{R}} &= \bm{\tilde{\mu}}^\top \left( \vec{q}{B}{R} \cdot \left( \vec{v}{IB}{B} + (\bm{\bar{\omega}} - \vec{b}{g}{})^{\times} \vec{r}{BR}{B} \right) \right)^{\times}.
\end{align}
Here, it is assumed that the noise from the radial speed measurement dominates over contributions from the gyroscope and bearing vector. Furthermore, correlations between the prediction and measurement steps, arising from the use of the gyroscope measurement in both, are neglected. This is reasonable due to the generally low noise magnitude in gyroscope measurements, whose influence is further reduced by the short lever arm between the radar and \ac{imu}. For outlier rejection, the Mahalanobis distance of the innovation term is evaluated against a Chi-squared distribution with one \ac{dof}.

\subsection{Feature Depth Initialization}
While the sparsity of the point cloud geometry limits its potential for inertial navigation aiding, it can still be used for feature initialization. For monocular vision, the feature depth is only observable over time, subject to parallax-induced changes in observability as a function of ego-motion. Furthermore, initialization often depends on heuristics, e.g., the median feature depth~\cite{bloesch2017Rovio}. This is problematic for short-lived features, environments with non-homogeneous depth, or non-dynamic motions. Thus, the radar range measurements are used to augment the feature depth initialization.

\begin{figure}[b!]
    \centering
    \vspace{-4ex}
    \includegraphics[width=\linewidth]{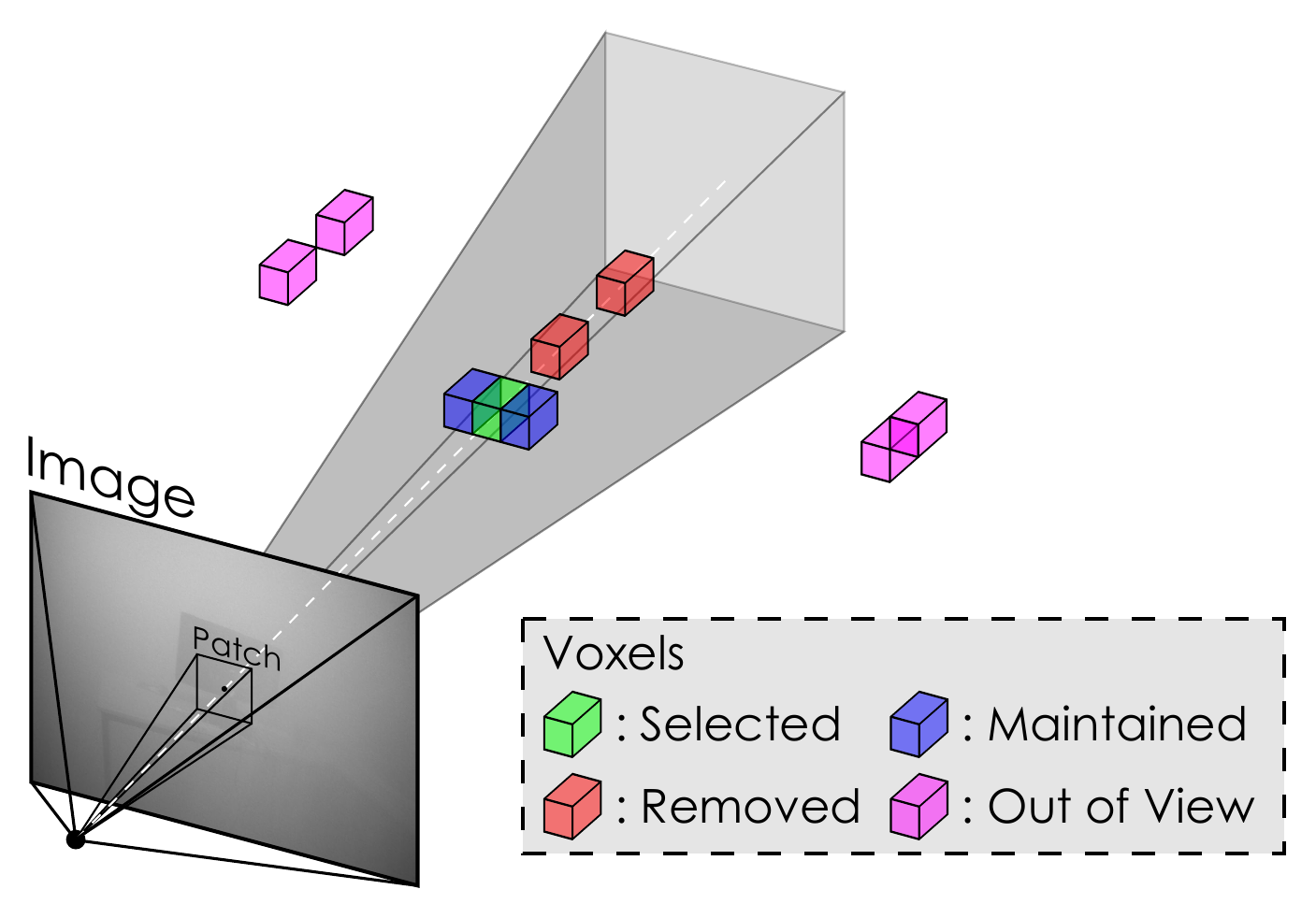}
    \spacingFigureCaption{}
    \caption{Correlation between the feature patch in the image and the voxels resulting from querying the feature depth.}
    \label{fig:depth_initialization}
    \spacingFigureText{}
\end{figure}

This requires filtering and aggregating radar points to address the noise and sparsity of the point clouds. To do so, a sliding-window voxel map with maximum range $\kappa_{d_{\max}}$ is created to maintain computational efficiency. Inlier radar measurements from the \ac{iekf} measurement update step are inserted into the voxel map, maintaining a minimum spacing between points $\kappa_{\Delta d}$
and a maximum number of points $\kappa_{N_{\max}}$ within a given voxel.

Upon initializing new image features, the voxels are iterated through, and their corners are projected into the image frame to determine, first, whether any part of the voxel lies in the image, and second, whether any part of the voxel lies in the feature patch. For each patch, the closest voxel with at least $\kappa_{N_{\min}}$ points is chosen, and its mean range is calculated for initialization. Candidate voxels with a distance greater than the selected voxel by at least the voxel size are then erased from the map as they are deemed obscured and likely consist of noisy measurements. An exemplary case of depth initialization is shown in \cref{fig:depth_initialization}, where multiple filled voxels are observed within a given feature patch's frustum, as expected from the properties of \ac{fmcw} radar sensing. Features without valid candidate voxels resort to the median-depth-based initialization routine.

\section{Evaluation}\label{sec:evaluation}
To evaluate the proposed method, flight experiments were conducted within a local forest (referred to as \forest{}), inside a university (referred to as \indoors{}), and above an open field (referred to as \field{}).

The experiments were conducted with a custom aerial robot platform equipped with a VectorNav VN100 \ac{imu}, an Ouster OS0-128 LiDAR, a Teledyne Blackfly S 0.4 MP camera, a Texas Instruments IWR6843AOPEVM radar, and a uBlox-M10S L1 \ac{gnss} sensor. The platform uses a microcontroller to synchronize the onboard sensing and computer timestamps following the approach from \cite{nissov2025Sync},
and the radar chirp is configured according to \cref{tab:evaluation:chirp}, following~\cite{doer2021yaw}. For this radar sensor and configuration, \qty{0.25}{\meter} voxels and values of $\kappa_{d_{\max}}=\SI{20}{\meter}$, $\kappa_{\Delta d}=\SI{0.05}{\meter}$, $\kappa_{N_{\max}}=20$, and $\kappa_{N_{\min}}=5$ are found to perform well.
Extrinsic transformations for all of the sensors are available from CAD, and the intrinsic and extrinsic parameters of the monocular camera are calibrated following \cite{rehder2016Kalibr}. Due to the quality of the camera-\ac{imu} calibration, the camera extrinsic estimation is disabled, although the framework exists for when the camera-\ac{imu} transformation is less well-known.

\begin{table}[h]
    \centering
    \caption{Radar Chirp Configuration}
    \label{tab:evaluation:chirp}
    \spacingTableCaption{}
    \newcommand{\rotation}{90}
\begin{tabular}{llS[table-format=2.3]l}
    \toprule
    \multicolumn{2}{l}{Parameter}  &{Value}   &Unit\\
    \midrule
        &Starting Frequency              &60 &\si{\giga\hertz}\\
    \midrule
    \multirow{2}{*}{\rotatebox{\rotation}{Max.}}
        &Range                   &20.013    &\si{\meter}\\
        &Doppler                 &3.995     &\si{\meter\per\second}\\
    \midrule
    \multirow{3}{*}{\rotatebox{\rotation}{Res.}}
        &Range                &0.078    &\si{\meter}\\
        &Doppler              &0.133    &\si{\meter\per\second}\\
        &Azimuth/Elevation    &29       &\si{\degree}\\
    \bottomrule
\end{tabular}
    \spacingTableText{}
\end{table}

The LiDAR and \ac{gnss} sensors are not used in the proposed method; however, they will be used to generate ground truth pose estimates (denoted \texttt{GT}). The LiDAR ground truth is based on the \ac{lio} solution presented in \cite{nissov2024degradation} (i.e., with the radar factors disabled) and is used in the \forest{} and \indoors{} environments. Here, high performance is expected due to the structured nature of these environments. The \ac{gnss} ground truth is created by fusing \ac{gnss} with \ac{imu} and barometer in a factor graph using a Levenberg-Marquardt optimizer. The \ac{gnss} ground truth is used in the \field{} environment, as here the geometry of the environment is not conducive to accurate \ac{lio}, following results in barren fields from~\cite{hatleskog2024Probabilistic,nissov2024robust}.

With respect to metrics, this work will evaluate the proposed method using the translation part of the \ac{ape} and \ac{rpe} (with \SI{10}{\meter} segment length)~\cite{grupp2017evo}, as well as the final position relative drift. The latter is calculated as the error of the final position estimate, normalized by the trajectory length.
This enables quantitative evaluation in environments where ground truth is otherwise not possible (e.g., with obscurants such as fog). For this, the platform's starting location is marked, and the platform is manually repositioned at the end of the experiment, careful to avoid obscuring exteroceptive sensors. As a result, the initial and final position estimates coincide (with an accuracy of approximately \SI{\pm 2}{\centi\meter}). For the \ac{ape}, the ground truth and estimated trajectories are aligned only by their initial poses.

In the following sections, we evaluate the proposed method (denoted \method{}), the proposed method with radar measurements disabled (denoted \vio{}), and the proposed method with image measurements disabled (denoted \rio{}). Note that the \vio{} configuration corresponds to ROVIO~\cite{bloesch2017Rovio}. Comparisons are also made against \rrxio{}~\cite{doer2021Rxio}, a state-of-the-art method for radar-visual fusion. All methods use the same parameters and initial estimates, with \ac{imu} noise parameters calculated from the Allan variance, except that \rrxio{} uses the covariance matrix estimated while calculating the linear velocity estimate with \ac{odr}.

The overall numerical results are shown in \cref{tab:results}.
In this, we denote instances when \ac{ape} and \ac{rpe} are not calculable due to a lack of ground truth by $-$ and instances where a given method outright fails by ${\times}$.

\begin{table*}[h]
    \centering
    \caption{Numerical Results of the Proposed Method across \Forest{}, \Indoors{}, and \Field{} Environments}
    \label{tab:results}
    \spacingTableCaption{}
    \newcommand{\rotation}[1]{\rotatebox{90}{#1}}
\scriptsize
\sisetup{
    tight-spacing           = true,
    table-text-alignment    = center,
    table-format            = 1.3
}
\newcommand{\spacing}{\hspace{3mm}}
\newcommand{\fail}{${\times}$}
\newcommand{\na}{${-}$}
\newcommand{\vmove}{-0.675ex}
\begin{tabular}{lS[round-precision=0,table-format=4.0] S@{\spacing}S@{\spacing}S@{\spacing}S S@{\spacing}S@{\spacing}S@{\spacing}S S@{\spacing}S@{\spacing}S@{\spacing}S}
    \toprule
    \multirow{2}{*}[\vmove]{Sequence}    &{\multirow{2}{*}[\vmove]{Length [\si{\meter}]}} &\multicolumn{4}{c}{\ac{ape} [\si{\meter}]}  &\multicolumn{4}{c}{\ac{rpe} ($\Delta:\SI{10}{\meter}$) [\si{\meter}]}  &\multicolumn{4}{c}{Final Position Relative Drift [\si{\centi\meter\per\meter}]}\\
    \cmidrule(lr){3-6}\cmidrule(lr){7-10}\cmidrule(lr){11-14}
        &   &\rio{}   &\vio{}   &\method{}   &\rrxio{}\cite{doer2021Rxio} &\rio{}   &\vio{}   &\method{} &\rrxio{}\cite{doer2021Rxio}   &\rio{}   &\vio{}   &\method{}   &\rrxio{}\cite{doer2021Rxio}\\
    \midrule
        \texttt{forest1}   &190   &2.311  &\textbf{0.722}  &0.744  &1.335    &0.128  &0.214  &\textbf{0.068}  &0.105    &0.506  &0.312  &0.311  &\textbf{0.187}\\
        \texttt{forest2}   &195   &3.298  &0.861  &\textbf{0.631}  &0.700    &0.193  &0.161  &\textbf{0.089}  &0.151    &1.030  &0.277  &\textbf{0.224}  &0.256\\
        \texttt{forest3}   &208   &2.727  &1.923  &\textbf{1.378}  &2.186    &0.178  &0.513  &0.080  &\textbf{0.079}    &0.640  &0.847  &\textbf{0.566}  &0.900\\
        \texttt{forest4}   &213   &2.618  &1.844  &\textbf{1.568}  &3.622    &0.201  &0.457  &0.135  &\textbf{0.118}    &0.891  &\textbf{0.415}  &0.434  &0.921\\
    \midrule
        \texttt{indoor\_dark1}  &210    &\textbf{0.371}  &0.550  &0.476  &0.700  &0.055  &0.112  &\textbf{0.045}  &0.118  &0.205  &0.227  &0.220  &\textbf{0.112}\\
        \texttt{indoor\_dark2}  &202    &\textbf{0.423}  &\fail  &0.497  &\fail  &0.056  &\fail  &\textbf{0.049}  &\fail  &\textbf{0.212}  &\fail  &0.339  &\fail\\
        \texttt{indoor\_fog1}   &65     &\na    &\na    &\na    &\na    &\na    &\na    &\na    &\na    &\textbf{0.276}  &\fail  &0.370  &\fail\\
        \texttt{indoor\_fog2}   &65     &\na    &\na    &\na    &\na    &\na    &\na    &\na    &\na    &\textbf{0.309}  &\fail  &0.394  &\fail\\
    \midrule
        \texttt{field1}   &1263   &\fail    &3.931  &\textbf{1.834}  &3.751    &\fail    &0.340  &\textbf{0.291}  &0.312    &\fail    &0.242  &\textbf{0.097}  &0.209\\
        \texttt{field2}   &1264   &\fail    &1.754  &1.859  &\textbf{1.423}    &\fail    &0.259  &\textbf{0.221}  &0.245    &\fail    &0.061  &0.080  &\textbf{0.037}\\
    \bottomrule
    \multicolumn{14}{l}{Instances where a metric cannot be computed are denoted by $-$ and instances of method failure are denoted by $\times$.}
\end{tabular}
    \spacingTableText{}
\end{table*}

\subsection{Nominal Conditions}
\begin{figure}[h]
    \centering
    \includegraphics[width=\linewidth]{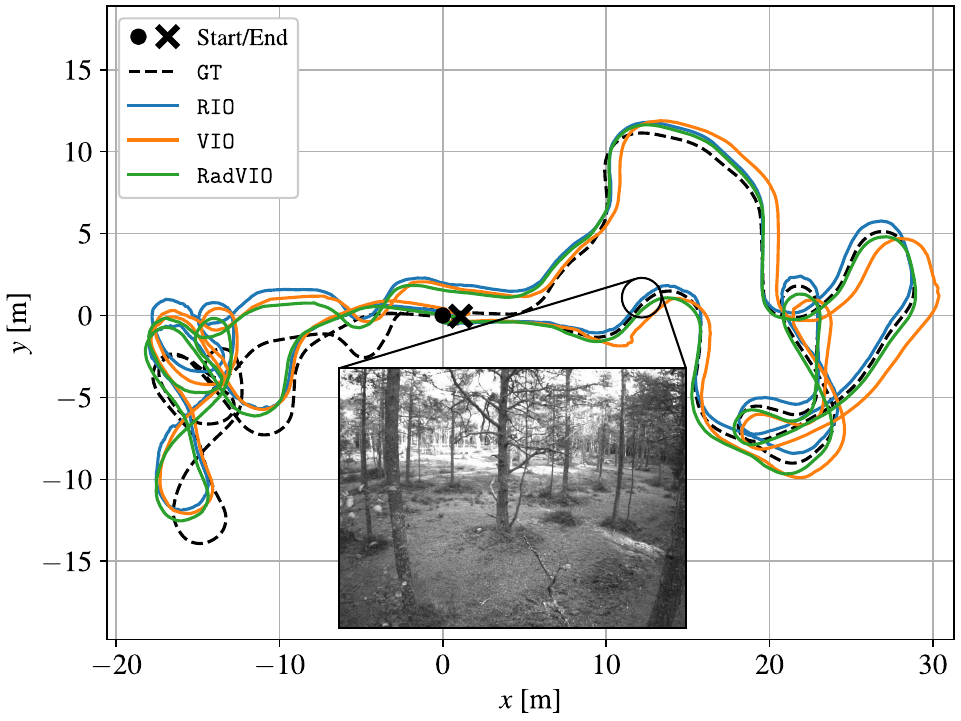}
    \spacingFigureCaption{}
    \caption{Position estimates of the proposed method alongside ablations from \experiment{forest4}. The image shows the good visual conditions of the \forest{} environment.}
    \label{fig:xy:forest}
    \spacingFigureText{}
\end{figure}

The \forest{} environment is well-suited to high performance for either \vio{} or \rio{} methods, as it provides ample objects for meaningful visual or radar measurements. However, the tight turns and high angular rates can make tracking visual features difficult. That said, both \rio{} and \vio{} perform well, as shown in \cref{tab:results}, and for \experiment{forest4} in \cref{fig:xy:forest}. Here, typical behavior of both odometries can be seen: highly accurate, local odometry from \rio{} (demonstrated by the \ac{rpe} scores) with improved long-term accuracy of \vio{} (demonstrated by the \ac{ape} scores). Despite both performing well in their own right, the overall performance improves when combined, as shown by \method{} in \cref{tab:results}. This improvement is particularly noticeable in the \ac{rpe} metric; for \ac{ape} it can be seen to vary. This is caused by the vertical and yaw drift associated with radar-based estimators~\cite{nissov2024degradation,noh2025garlio,doer2020ekf}, which cannot be entirely mitigated by vision due to the known lack of observability~\cite{jones2011VIO}.

\begin{figure}[h]
    \centering
    \includegraphics[width=\linewidth]{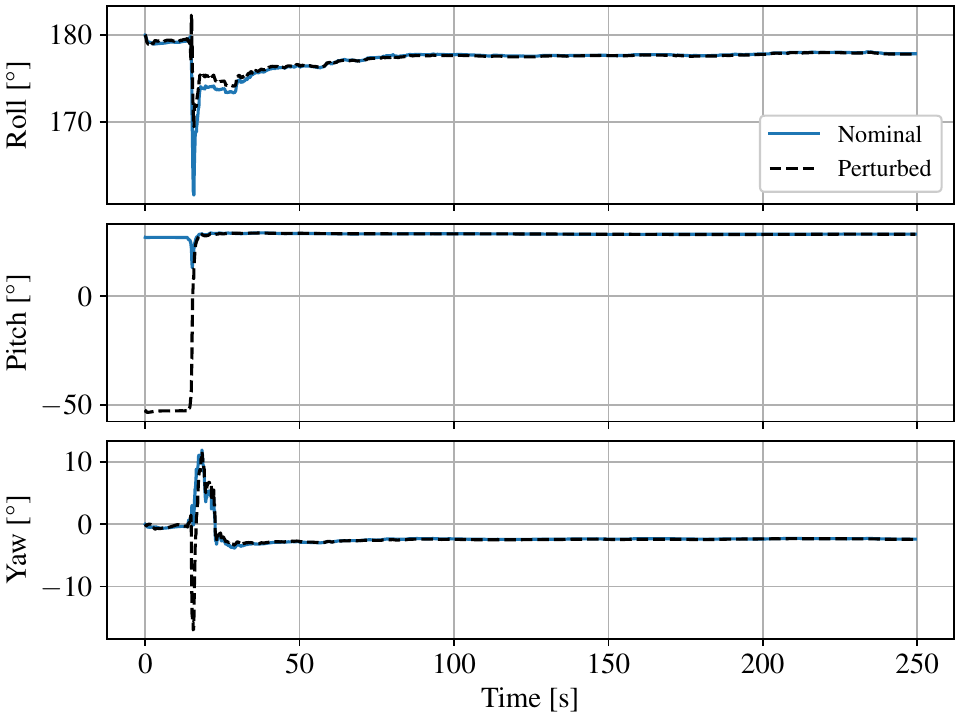}
    \spacingFigureCaption{}
    \caption{Radar extrinsic estimation of \method{} in \experiment{forest1} with nominal and perturbed initial guesses. The perturbed initial guess is generated by applying a rotation of \SI{80}{\degree} about the radar sensor's $y$ axis to the nominal prior.}
    \label{fig:xy:extrinsics}
    \spacingFigureText{}
\end{figure}

An additional benefit of the proposed method is the ability to refine the extrinsic calibration online. Where established routines exist for calibrating the extrinsic transform between camera and \ac{imu} sensors~\cite{rehder2016Kalibr}, the same is not as common for radars. While an initial guess can typically be provided with high accuracy, even small rotation offsets can result in significant drift over time. This is because the radar measurement update directly informs velocity rather than position. This effect is visible in the performance of \rrxio{} in this environment, with slight extrinsic error resulting in increased \ac{ape}. Thus, for the radar, online extrinsic estimation is important for refining the initial guess. To exaggerate this, \method{} is re-run on \experiment{forest1} with an inaccurate initial estimate of the radar-\ac{imu} extrinsic. This is made by perturbing the rotation prior by \SI{80}{\degree} about the radar $y$ axis. Compared with the nominal prior, as shown in \cref{fig:xy:extrinsics}, the proposed method is able to quickly converge to the same value despite high initial error.

\subsection{Challenging Environments for Vision}
To demonstrate the main benefits of the multi-modal fusion, robustness in challenging conditions, experiments were conducted specifically targeting typical weaknesses of \vio{}: challenging lighting conditions and obscurants.

Note that in the indoor environment, we observe that \rrxio{} has difficulty accurately estimating the linear velocity and covariance matrix, to the point that if the image measurements were disabled, the solution would diverge. As a result, one can see it fail in circumstances that might seem surprising, such as in darkness or fog. This is because the environment effectively disables the image feature updates, and the radar portion of the estimator is underperforming.

\subsubsection{Darkness}\label{sec:evaluation:darkness}
\begin{figure}[h]
    \centering
    \includegraphics[width=\linewidth]{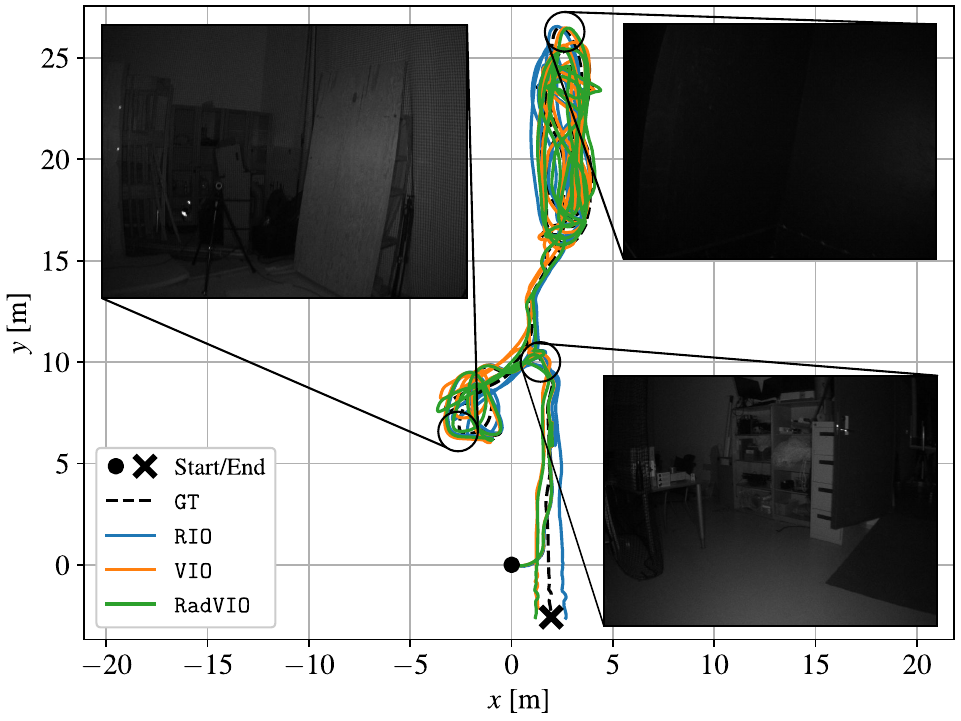}
    \spacingFigureCaption{}
    \caption{Position estimates of the proposed method alongside ablations from \experiment{indoor\_dark1}. Even with onboard lighting, the image visibility, and therefore the availability of good features, varies across the environment.}
    \label{fig:xy:darkness}
    \spacingFigureText{}
\end{figure}

The aerial platform was flown manually through a dark environment, both with (in \experiment{indoor\_dark1}) and without (in \experiment{indoor\_dark2}) any onboard lighting. Onboard lighting illuminates the environment near the platform; however, objects further away may still be difficult to observe. 
The performance of both experiments is shown in \cref{tab:results}, where darkness, even with onboard lighting, affects the vision-based method's performance. Without onboard lighting, the \vio{} estimator cannot extract features and fails entirely. In both cases, \rio{} performs well, and further improvements in performance are observed when both exteroceptive modalities are combined in \method{}. The results for \experiment{indoor\_dark1} are shown in \cref{fig:xy:darkness}, with images from different regions of the environment to illustrate the changing lighting conditions.

\subsubsection{Fog}
\begin{figure}[h]
    \centering
    \includegraphics[width=\linewidth]{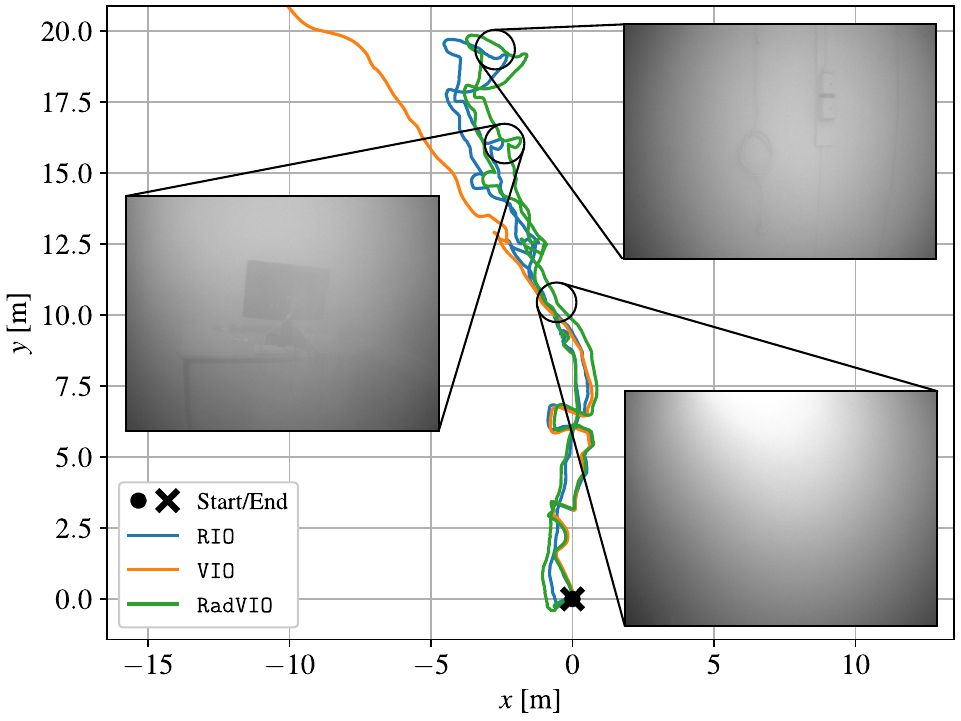}
    \spacingFigureCaption{}
    \caption{Position estimates of the proposed method alongside ablations from \experiment{indoor\_fog2}. One can clearly see how, in some regions, objects can be made out, whereas in others the image is rendered useless.}
    \label{fig:xy:fog}
    \spacingFigureText{}
\end{figure}

The aerial platform was flown through an indoor environment (different from \cref{sec:evaluation:darkness}), which was filled with dense fog. As a result, the camera images, in the worst case, are completely devoid of features due to the substantially degraded visibility. Occasionally, a few features will be visible; however, such instances are too sparse for reliable odometry, resulting in \vio{} diverging quickly. \rio{}, on the other hand, does not suffer the same challenge due to the radar's longer wavelength, and as a result, both \rio{} and \method{} return approximately to the starting location. As LiDAR sensors also show performance issues in such environments~\cite{nissov2024degradation}, sourcing ground truth from \ac{lio} will not be possible. As a result, we rely on the relative drift metric. Given the lack of observability~\cite{jones2011VIO}, this is not an ideal metric as noise from short-lived features in the final, singular pose estimate can impact the evaluation of the entire trajectory. With that in mind, the negative impacts of including short-lived, noisy features can be seen as the performance of \method{} is slightly degraded. For prolonged flight in such environments, performance could be improved by restricting feature selection.
Position estimates for the \experiment{indoor\_fog2} experiment are shown in \cref{fig:xy:fog}, along with representative images of different regions of the environment.

\subsection{Challenging Environments for Radar}
In similar works investigating multi-modal fusion between \ac{fmcw} radar and another exteroceptive modality~\cite{nissov2024degradation,nissov2024robust,noh2025garlio,doer2021Rxio}, the radar is typically seen as the infallible, but noisy, counterpart to an otherwise failing vision or LiDAR. Scenarios where the \ac{fmcw} radar is ill-posed are not so commonly considered. Aiming to overcome this gap, two challenges of \ac{fmcw} radar sensors are examined in this work: the restriction on maximum Doppler due to the chirp design and even sparser measurements resulting from environments with poor reflectivity.

The characteristics of the radar chirp result in resolution and maximum limits for the measurements, where the maximum is related to the Nyquist theorem. Similar to exposure time, an improper selection of chirp parameters can lead to suboptimal results. However, a key difference here is that the chirp configuration is typically not changeable at runtime and is limited by the physical characteristics of the sensor. As a result, conducting high-speed flights with a radar sensor improperly configured or outside its design envelope can lead to poor-quality measurements.
Furthermore, environments with poor reflectivity will yield few, valid measurements due to limitations with respect to \ac{snr}. Comparisons could be drawn to cameras not optimized for low-light performance: even if adjusted online, the intersection of sufficient exposure without excessive motion blur may not exist for a given sensor under certain trajectories in difficult environments.

\begin{figure}[h]
    \centering
    \includegraphics[width=\linewidth]{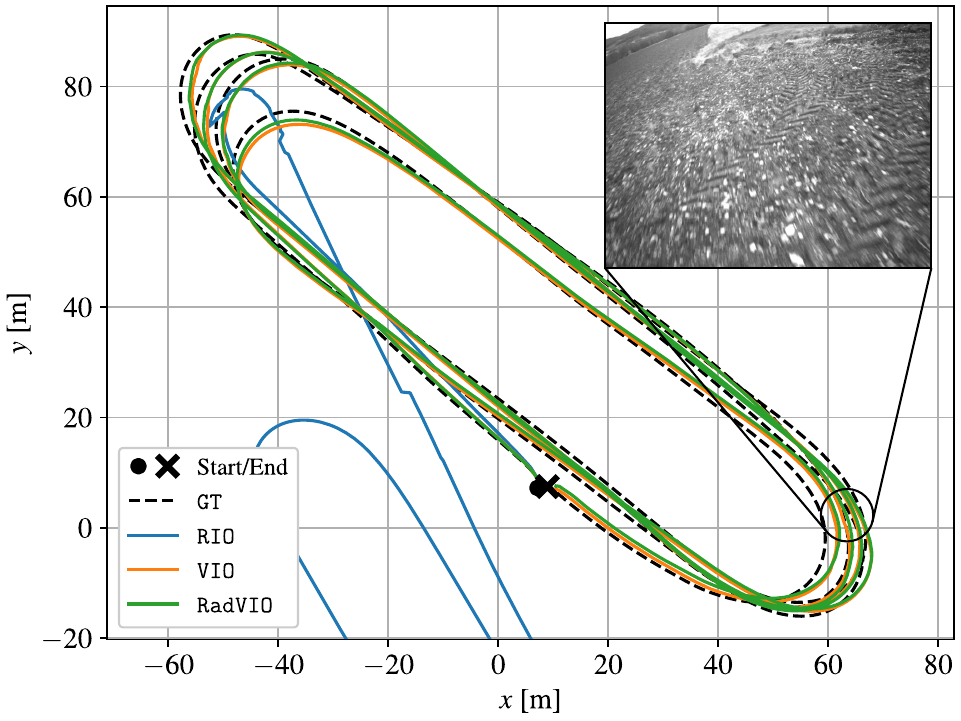}
    \spacingFigureCaption{}
    \caption{Position estimates of the proposed method alongside ablations from \experiment{field2}. The degraded radar performance is clear, resulting from limited clutter and high speeds (up to \SI{11}{\meter\per\second}).}
    \label{fig:xy:field}
    \spacingFigureText{}
\end{figure}

\begin{figure}[h]
    \centering
    \includegraphics[width=\linewidth]{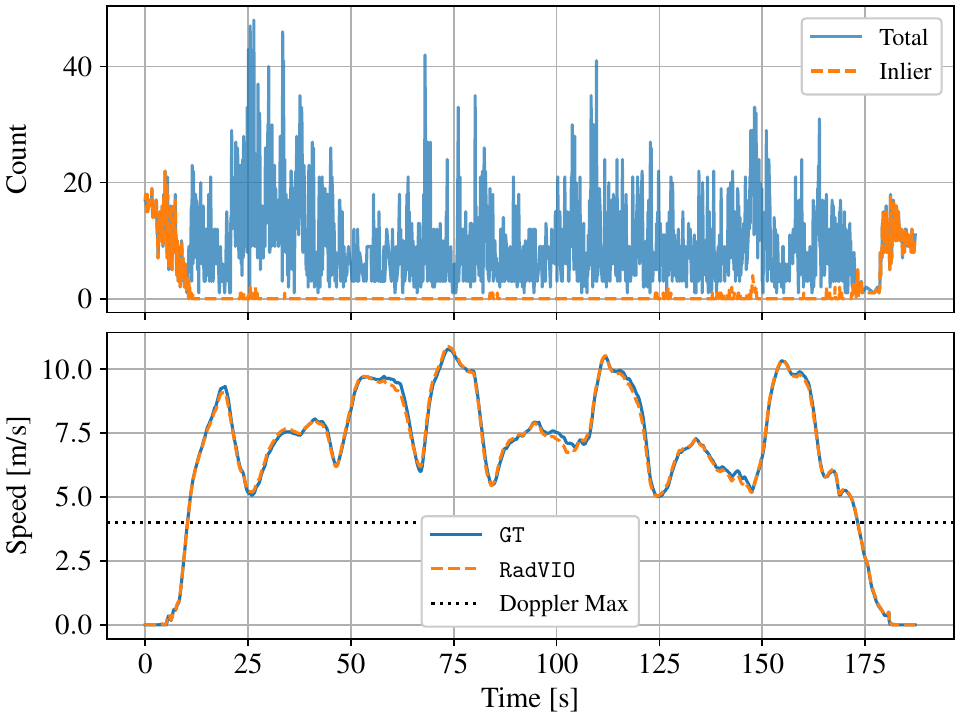}
    \spacingFigureCaption{}
    \caption{The radar point cloud size, number of inliers from \method{} measurement updates, and velocity estimates, all from \experiment{field2}.}
    \label{fig:field:cloud_size}
    \spacingFigureText{}
\end{figure}

To provoke both challenges simultaneously, the aerial platform is flown at high speed (up to \SI{11.4}{\meter\per\second} in \experiment{field1}) above an empty field, reducing radar returns by removing infrastructure from the environment. The result, as shown in \cref{tab:results}, is that \rio{} alone cannot perform. This is evident from the diverging estimates of the \experiment{field2} experiment shown in \cref{fig:xy:field}. Periods with low numbers of inlier measurements, shown in \cref{fig:field:cloud_size} for the same experiment, cause the filter to diverge or drift uncontrollably. However, the measurements themselves are still usable, as when paired with the vision updates, \method{} can hold over between periods of radar outages. This is a distinct advantage of fusing radial speed, as when estimating the full linear velocity, one cannot easily recover the partial information. Naturally, the \ac{odr} estimate used by \rrxio{} is compromised by the high number of outliers during the high-speed sections of the trajectory.

\section{Conclusions}\label{sec:conclusions}
This manuscript investigated the multi-modal fusion of vision and \ac{fmcw} radar for aided inertial navigation, leveraging the strengths of each exteroceptive modality to mitigate their shortcomings and achieve greater robustness. A tightly-coupled estimator was proposed to fuse radial speed measurements from radar and feature-intensity updates from vision, while using the radar range for image feature depth initialization. The proposed method was evaluated across three environments, specifically designed to highlight key failure modes of radar- and camera-based methods. The proposed method demonstrated superior performance, even when both modalities saw favorable conditions. Furthermore, performance remained consistent in environments with fog and darkness (challenging for vision) and at excessively high speeds (challenging for radar), where each \ac{vio} and \ac{rio} alone fail, respectively.

\raggedbottom 
\bibliographystyle{IEEEtran}
\bibliography{bib/general}

\end{document}